\documentclass{article} 
\usepackage{iclr2021_conference,times}

\usepackage{hyperref}
\usepackage{url}

\usepackage{subfig}
\usepackage{multirow, multicol}
\usepackage{float}
\usepackage{graphicx}
\usepackage{multirow, multicol}
\usepackage[colorinlistoftodos,prependcaption]{todonotes}
\usepackage{tabulary,booktabs}
\usepackage{tablefootnote}
\usepackage{amsfonts}
\usepackage{color,soul}
\usepackage{multirow}
\usepackage{todonotes}
\usepackage{amsmath}
\usepackage{colortbl}
\usepackage{color, xcolor}
\usepackage{bm}
\usepackage[subtle]{savetrees}
\usepackage{comment}

\newcommand{\shortpaper}[1]{}
\newcommand{\remove}[1]{}
\newcommand{\removed}[1]{} 
\newcommand{\checkandremove}[1]{} 
\DeclareMathOperator*{\expect}{\mathbb{E}}
\usepackage{array}
\newcolumntype{H}{>{\setbox0=\hbox\bgroup}c<{\egroup}@{}}

\title{
Variational Information Bottleneck for Effective Low-Resource Fine-Tuning
}

\author{
Rabeeh Karimi Mahabadi$^{\clubsuit\heartsuit}$ \hspace{4em} Yonatan Belinkov$^{\diamondsuit}$\thanks{Supported by the Viterbi Fellowship in the Center for Computer Engineering at the Technion.} \hspace{4em} James Henderson$^\clubsuit$\\
$^\heartsuit$EPFL, Switzerland\\
$^\clubsuit$Idiap Research Institute, Switzerland \\
$^\diamondsuit$Technion -- Israel Institute of Technology\\
{ \tt \{rabeeh.karimi,james.henderson\}@idiap.ch} \\ 
{ \tt belinkov@technion.ac.il}
}

\iclrfinalcopy 
\begin{document}

\maketitle
\begin{abstract}
While large-scale pretrained language models have obtained impressive results when fine-tuned on a wide variety of tasks, they still often suffer from overfitting in low-resource scenarios. Since such models are general-purpose feature extractors, many of these features are inevitably irrelevant for a given target task.  We propose to use Variational Information Bottleneck (VIB) to suppress irrelevant features when fine-tuning on low-resource target tasks, and show that our method successfully reduces overfitting.  Moreover, we show that our VIB model finds sentence representations that are more robust to biases in natural language inference datasets, and thereby obtains better generalization to out-of-domain datasets. Evaluation on seven low-resource datasets in different tasks shows that our method significantly improves transfer learning in low-resource scenarios, surpassing prior work. Moreover, it improves generalization on 13 out of 15 out-of-domain natural language inference benchmarks.  Our code is publicly available in \url{https://github.com/rabeehk/vibert}.
\end{abstract}

\section{Introduction}
Transfer learning has emerged as the de facto standard technique in natural language processing (NLP), where large-scale language models are pretrained on an immense amount of text to learn a general-purpose representation, which is then transferred to the target domain with fine-tuning on target task data. This method has exhibited state-of-the-art results on a wide range of NLP benchmarks~\citep{devlin2019bert, liu2019roberta, radford2019language}.
However, such pretrained models have a huge number of parameters, potentially making fine-tuning susceptible to overfitting.

In particular, the task-universal nature of large-scale pretrained sentence representations means that much of the information in these representations is irrelevant to a given target task.  If the amount of target task data is small, it can be hard for fine-tuning to distinguish relevant from irrelevant information, leading to overfitting on statistically spurious correlations between the irrelevant information and target labels.
Learning low-resource tasks is an important topic in NLP~\citep{emnlp-2019-deep} because annotating more data can be very costly and time-consuming, and because in several tasks access to data is limited.

In this paper, we propose to use the Information Bottleneck (IB) principle~\citep{Tishby99theinformation} to address this problem of overfitting.  More specifically, we propose a fine-tuning method that uses Variational Information Bottleneck (VIB;~\citealt{alemi2016deep}) to improve transfer learning in low-resource scenarios.

\begin{figure}[t]\vspace{-1em}
\centerline{
\includegraphics[width=0.9\textwidth]{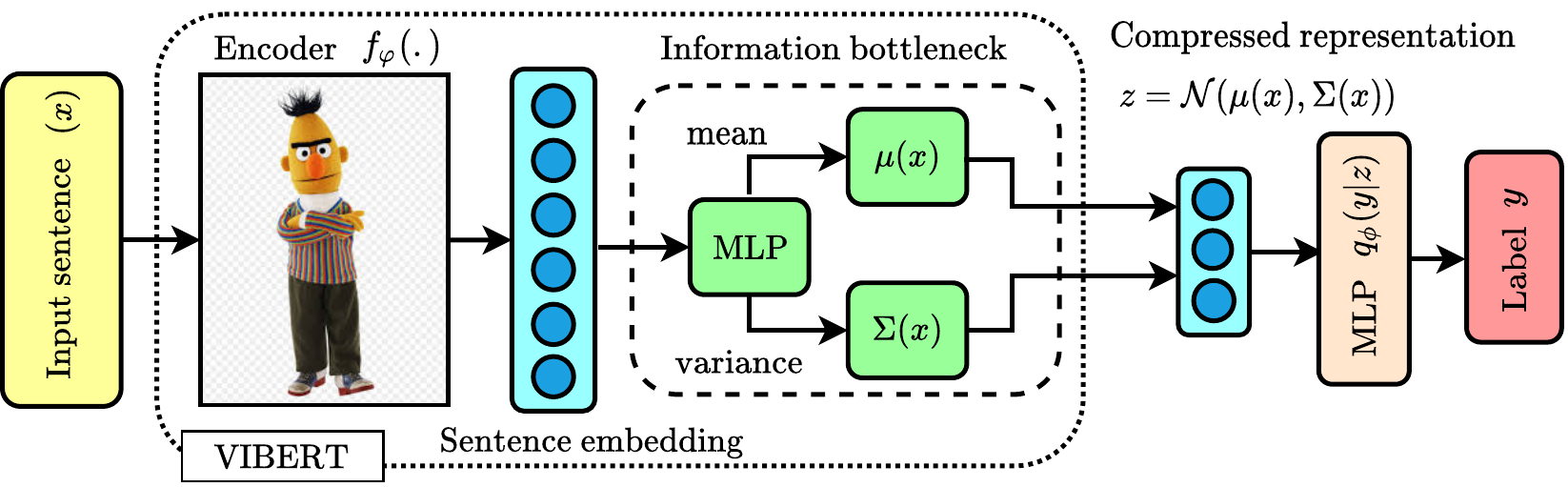}}
\vspace{-1em}
\caption{VIBERT compresses the encoder's sentence representation $f_\varphi(x)$ into representation $z$ with mean $\mu(x)$ and eliminates irrelevant and redundant information through the Gaussian noise with variance $\Sigma(x)$.}
\vspace{-1.4em}
\label{fig:IB}
\end{figure}

VIB addresses the problem of overfitting by adding a regularization term to the training loss that directly suppresses irrelevant information.   As illustrated in Figure~\ref{fig:IB}, the VIB component maps the sentence embedding from the pretrained model to a latent representation $z$, which is the only input to the task-specific classifier. The information that is represented in $z$ is chosen based on the IB principle, namely that all the information about the input that is represented in $z$ should be necessary for the task.  In particular, VIB directly tries to remove the irrelevant information, making it easier for the task classifier to avoid overfitting when trained on a small amount of data.
We find that in low-resource scenarios, using VIB to suppress irrelevant features in pretrained sentence representations substantially improves accuracy on the target task.  

Removing unnecessary information from the sentence representation also implies removing redundant information. VIB tries to find the most concise representation which can still solve the task, so even if a feature is useful alone, it may be removed if it isn't useful when added to other features because it is redundant. We hypothesize that this provides a useful inductive bias for some tasks, resulting in better generalization to out-of-domain data.  In particular, it has recently been demonstrated that annotation biases and artifacts in several natural language understanding benchmarks~\citep{kaushik2018much, gururangan2018annotation, poliak2018hypothesis, schuster2019towards} allow models to exploit superficial shortcuts during training to perform surprisingly well without learning the underlying task.  However, models that rely on such superficial features do not generalize well to out-of-domain datasets, which do not share the same shortcuts~\citep{belinkov-etal-2019-dont1}.
We investigate whether using VIB to suppress redundant features in pretrained sentence embeddings has the effect of removing these superficial shortcuts and keeping the deep semantic features that are truly useful for learning the underlying task.
We find that using VIB does reduce the model's dependence on shortcut features and substantially improves generalization to out-of-domain datasets.  

We evaluate the effectiveness of our method on fine-tuning BERT~\citep{devlin2019bert}, which we call the VIBERT model (Variational Information Bottleneck for Effective Low-Resource Fine-Tuning).  On seven different datasets for text classification, natural language inference, similarity, and paraphrase tasks, VIBERT shows greater robustness to overfitting than conventional fine-tuning and other regularization techniques, improving accuracies on low-resource datasets.  Moreover, on NLI datasets, VIBERT shows robustness to dataset biases, obtaining substantially better generalization to out-of-domain NLI datasets.  
Further analysis demonstrates that VIB regularization results in less biased representations.
Our approach is highly effective and simple to implement, involving a small additional MLP classifier on top of the sentence embeddings. It is model agnostic and end-to-end trainable.

In summary, we make the following contributions:
1)~Proposing VIB for low-resource fine-tuning of large pretrained language models.
2)~Showing empirically that VIB reduces overfitting, resulting in substantially improved accuracies on seven low-resource benchmark datasets against conventional fine-tuning and prior regularization techniques.
3)~Showing empirically that training with VIB is more robust to dataset biases in NLI, resulting in significantly improved generalization to out-of-domain NLI datasets.
To facilitate future work, we will release our code.

\section{Fine-tuning in Low-resource Settings}

The standard fine-tuning paradigm starts with a large-scale pretrained model such as BERT, adds a task-specific output component which uses the pretrained model's sentence representation, and trains this model end-to-end on the task data, fine-tuning the parameters of the pretrained model.  As depicted in Figure~\ref{fig:IB}, we propose to add a VIB component that controls the flow of information from the representations of the pretrained model to the output component.  The goal is to address overfitting in resource-limited scenarios by removing irrelevant and redundant information from the pretrained representation.

\vspace{-1ex}

\paragraph{Problem Formulation} We consider a general multi-class classification problem with a low-resource dataset $\mathcal{D}=\{x_i, y_i\}_{i=1}^N$ consisting of inputs $x_i \in \mathcal{X}$, and labels $y_i \in \mathcal{Y}$.  We assume we are also given a large-scale pretrained encoder $f_\varphi(.)$ parameterized by $\varphi$ that computes sentence embeddings  for the input $x_i$. Our goal is to fine-tune $f_\varphi(.)$ on $\mathcal{D}$ to maximize generalization.

\vspace{-1ex}
\paragraph{Information Bottleneck}
To specifically optimize for the removal of irrelevant and redundant information from the input representations, we adopt the Information Bottleneck principle. The objective of IB is to find a maximally compressed representation $Z$ of the input representation $X$ (compression loss) that maximally preserves information about the output $Y$ (prediction loss),\footnote{In this work, $Z$, $X$, and $Y$ are random variables, and $z$, $x$ and $y$ are instances of these random variables.} by minimizing:
\vspace{-0.5ex}
\begin{align}
\mathcal{L_\text{IB}} =\underbrace{\beta\, I(X,~Z)}_{\text{Compression Loss}}-\underbrace{I(Z,~Y)}_{\text{Prediction Loss}}, 
\label{eq:ib}
\\[-4.5ex]\nonumber
\end{align} 
where $\beta \geq 0$ controls the balance between compression and prediction, and $I(., .)$ is the mutual information.

\paragraph{Variational Information Bottleneck}\citet{alemi2016deep} derive an efficient variational estimate of~\eqref{eq:ib}:
\begin{align} 
\mathcal{L}_\text{VIB} =\beta~\expect_{x} \left[ \text{KL} \left[p_{\theta}(z|x),~r(z)\right]\right]+ \expect_{z\sim p_{\theta}(z|x)}[-\log q_{\phi}(y|z)],
\end{align} where $q_{\phi}(y|z)$ is a parametric approximation of $p(y|z)$, $r(z)$ is an estimate of the prior probability $p(z)$ of $z$, and $p_{\theta}(z|x)$ is an estimate of the posterior probability of $z$. During training, the compressed sentence representation $z$ is sampled from the distribution $p_{\theta}(z|x)$, meaning that a specific pattern of noise is added to the input of the output classifier $q_{\phi}(y|z)$. Increasing this noise decreases the information conveyed by $z$.  In this way, the VIB module can block the output classifier $q_{\phi}(y|z)$ from learning to use specific information.
At test time, the expected value of $z$ is used for predicting labels with $q_{\phi}(y|z)$. We refer to the dimensionality of $z$ as $K$, which specifies the bottleneck size.
Note that there is an interaction between decreasing $K$ and increasing the compression by increasing $\beta$~\citep{shamir2010learning,harremoes2007information}. $K$ and $\beta$ are hyper-parameters~\citep{alemi2016deep}.

We consider parametric Gaussian distributions for prior $r(z)$ and $p_{\theta}(z|x)$ to allow an analytic computation for their Kullback-Leibler divergence,\footnote{$\text{KL}(\mathcal{N}(\mu_0, \Sigma_0)\|\mathcal{N}(\mu_1, \Sigma_1)) = \frac{1}{2}(\text{tr}(\Sigma_1^{-1}\Sigma_0)+(\mu_1-\mu_0)^T\Sigma_1^{-1}(\mu_1-\mu_0)-K+\log(\frac{\det(\Sigma_1)}{\det(\Sigma_0)}))$.} namely $r(z) = \mathcal{N}(z|\mu_0, \Sigma_0)$ and
$p_{\theta}(z|x) = \mathcal{N}(z |\mu(x), \Sigma(x))$, where $\mu$ and $\mu_0$ are $K-$dimensional mean vectors, and $\Sigma$ and $\Sigma_0$ are diagonal covariance matrices. We use the reparameterization trick~\citep{kingma2013auto} to estimate the gradients, namely $z=\mu(x)+\Sigma(x)\odot \epsilon$, where $\epsilon \sim \mathcal{N}(0,I)$. To compute the compressed sentence representations $p_{\theta}(z|x)$,  as shown in Figure~\ref{fig:IB},  we first feed sentence embeddings $f_\varphi(x)$ through a shallow MLP. It is then followed by two linear layers, each with $K$ hidden units to compute $\mu(x)$ and $\Sigma(x)$ (after a softplus transform to ensure non-negativity). We also use another linear layer to approximate $q_{\phi}(y|z)$.

\section{Experiments}
\paragraph{Datasets} We evaluate the performance on seven different benchmarks for multiple tasks, in particular text classification, natural language inference, similarity, and paraphrase detection. For NLI, we experiment with two well-known NLI benchmarks, namely SNLI~\citep{bowman2015large} and MNLI~\citep{williams2018broad}. For text classification, we evaluate on two sentiment analysis datasets, namely IMDB~\citep{maas2011learning} and Yelp2013 (YELP)~\citep{zhang2015character}. We additionally evaluate on three low-resource datasets in the GLUE benchmark~\citep{wang2018glue}:\footnote{We did not evaluate on WNLI and CoLA due to the irregularities in these datasets and the reported instability during the fine-tuning~\url{https://gluebenchmark.com/faq}.} 
paraphrase detection using MRPC~\citep{dolan2005automatically}, semantic textual similarity using STS-B~\citep{cer2017semeval}, and textual entailment using RTE~\citep{dagan2006pascal}. For the GLUE benchmark, SNLI, and Yelp, we evaluate on the standard validation and test splits. For MNLI, since the test sets are not available, we tune on the matched dev set and evaluate on the mismatched dev set (MNLI-M) or vice versa. See Appendix~\ref{app:experiments_details} for datasets statistics and Appendix~\ref{app:hyperparameters} for hyper-parameters of all methods. 
\paragraph{Base Model} We use the BERT$_\text{Base}$ (12 layers, 110M parameters) and BERT$_\text{Large}$ (24 layers, 340M parameters) uncased~\citep{devlin2019bert} implementation of~\citet{Wolf2019HuggingFacesTS} as our base models,\footnote{To have a controlled comparison, all results are computed with this PyTorch implementation, which might slightly differ from the TensorFlow variant~\citep{devlin2019bert}.} known to work well for these tasks. We use the default hyper-parameters of BERT, i.e., we use a sequence length of $128$, with batch size $32$. We use the stable variant of the Adam optimizer~\citep{zhang2020revisiting,mosbach2020stability} with the default learning rate of $2\mathrm{e}{-5}$ through all experiments. We do not use warm-up or weight decay.

\paragraph{Baselines} We compare against prior regularization techniques, including previous state-of-the-art, Mixout:
\begin{itemize}
  \item \textbf{Dropout}~\citep{srivastava2014dropout}, a widely used stochastic regularization techniques used in multiple large-scale language models~\citep{devlin2019bert, yang2019xlnet, vaswani2017attention} to mitigate overfitting. Following~\citet{devlin2019bert}, we apply dropout on all layers of BERT.
  
    \item \textbf{Mixout}~\citep{lee2019mixout} is a stochastic regularization technique inspired by Dropout with the goal of preventing catastrophic forgetting during fine-tuning. Mixout regularizes the learning to minimize the deviation of a fine-tuned model from the pretrained initialization. It replaces the model parameters with the corresponding value from the pretrained model with probability $p$.
    \item\textbf{Weight Decay~(WD)} is a common regularization technique to improve generalization~\citep{krogh1992simple}. It regularizes the large weights $w$ by adding a penalization term $\frac{\lambda}{2}\|w\|$ to the loss, where $\lambda$ is a hyperparameter specifying the strength of regularization. \citet{chelba2004adaptation} and~\citet{daume2007frustratingly} adapt WD for fine-tuning of the pretrained models, and propose to replace this regularization term with $\lambda \|w-w_0\|$, where $w_0$ are the weights of the pretrained models. Recently,~\citet{lee2019mixout} demonstrated that the latter formulation of WD works better for fine-tuning of BERT than conventional WD and can improve generalization on small training sets. 
\end{itemize}

\subsection{Results on the GLUE Benchmark}\label{sec:glue_results}
Table~\ref{tab:glue_results} shows results on the low-resource datasets in GLUE.\footnote{Note that the test sets are not publicly available and the prior work reports the results on the validation set of the GLUE benchmark~\citep{lee2019mixout, dodge2020fine}. We, however, report the results of their methods and ours on the original test sets by submitting to an online system.} We find that a) Our VIBERT model substantially outperforms the baselines on all the datasets, demonstrating the effectiveness of the proposed method. b) Dropout decreases the performance on low-resource datasets. We conjecture that regularization techniques relying on stochasticity without considering the relevance to the output, in contrast to VIB, can make it more difficult for learning to extract relevant information from a small amount of data. \citet{igl2019generalization} observe similar effects in another application. c) Similar to the results of~\cite{zhang2020revisiting}, we find less pronounced benefits of the previously suggested methods than the results originally published. This can be explained by using a more stable version of Adam~\citep{zhang2020revisiting} suggested by the very recent work in our experiments, which decreases the added benefits of previously suggested regularization techniques on top of a stable optimizer. In contrast, our VIBERT model still substantially improves the results and surpasses the prior work in all settings for both BERT$_\text{Base}$ and BERT$_\text{Large}$ models. 
Due to the computational overhead of BERT$_\text{Large}$, for the rest of this work, we stick to BERT$_\text{Base}$. 
\begin{table}[H] 
\vspace{-0.8em}
 \caption{Average results and standard deviation in parentheses over 3 runs on low-resource data in GLUE. $\bm{\Delta}$ shows the absolute difference between the results of the VIBERT model with BERT.}
    \centering \vspace{-0.5em}
    \resizebox{1.0\textwidth}{!}{
    \begin{tabular}{lllllll} 
        \toprule 
         & \multicolumn{2}{c}{\textbf{MRPC}} & \multicolumn{2}{c}{\textbf{STS-B}} & \multicolumn{1}{c}{\textbf{RTE}}\\ 
         \cmidrule(r){2-3} \cmidrule(l){4-5}  \cmidrule(l){6-6}
        {\vspace{-0.75em} \textbf{Model}} &  \multicolumn{2}{c}{} & \multicolumn{2}{c}{} &\multicolumn{1}{c}{} \\ 
                 & \textbf{Accuracy} & \textbf{F1}  &\textbf{Pearson} & \textbf{Spearman} & \textbf{Accuracy}\\  
    \toprule 
    BERT$_\text{Base}$ & 87.80 (0.5) & 83.20 (0.6) & 84.93 (0.1) & 83.53 (0.0) & 67.93 (1.5) \\ 
    ~~+Dropout~\citep{srivastava2014dropout} & 87.33 (0.2) & 81.90 (0.7) & 84.33 (0.9) & 82.73(1.0) & 65.80 (1.5) \\
    ~~+Mixout~\citep{lee2019mixout} & 87.03 (0.2) & 82.63 (0.3) & 85.23 (0.4)& 83.80(0.4) & 67.70 (0.9) \\ 
    ~~+WD~\citep{lee2019mixout} & 87.57(0.2) & 82.83(0.3) & 85.0(0.3) & 83.6(0.2) & 68.63(1.3) \\
    \midrule
    VIBERT$_\text{Base}$ & \textbf{89.23 (0.1)} & \textbf{85.23 (0.2)} & \textbf{87.63 (0.3)} & \textbf{86.50 (0.4)} & \textbf{70.53 (0.5)} \\
    $\bm{\Delta}$ & +1.43 & +2.03 & +2.7 & +2.97 & +2.6\\ 
    \toprule
     BERT$_\text{Large}$ &88.47 (0.7) & 84.20 (1.3) & 86.87 (0.2) & 85.70 (0.1) & 68.67 (0.8)\\ 
      ~~+Dropout~\citep{srivastava2014dropout}&  87.77 (0.4) & 82.97 (0.2) & 86.47 (0.1) &85.33 (0.2) & 65.77 (0.6) \\ 
      ~~+Mixout~\citep{lee2019mixout} & 88.57 (0.7) & 84.10 (1.1) & 86.70 (0.2) &85.43 (0.3) & 70.03 (1.0) \\
      ~~+WD~\citep{lee2019mixout} & 88.97(0.5) & 84.87(0.4) & 86.9(0.1) & 85.67(0.1) & 69.27(0.9) \\ 
     \midrule 
      VIBERT$_\text{Large}$ & \textbf{89.10 (0.4)} & \textbf{85.13 (0.6)} & \textbf{87.53 (0.8)} & \textbf{86.40 (0.9)} & \textbf{71.37 (0.8)} \\
    $\bm{\Delta}$ & +0.63 &  +0.93 &  +0.66 &  +0.7 & +2.7 \\ 
    \bottomrule
    \end{tabular}}
    \vspace{-0.5em}
\label{tab:glue_results}
    \vspace{-1em}
\end{table}

\begin{figure}[t!]
\centering \vspace{-2em}
\begin{minipage}[t]{.3\linewidth}
    \centering
    \includegraphics[width=\linewidth]{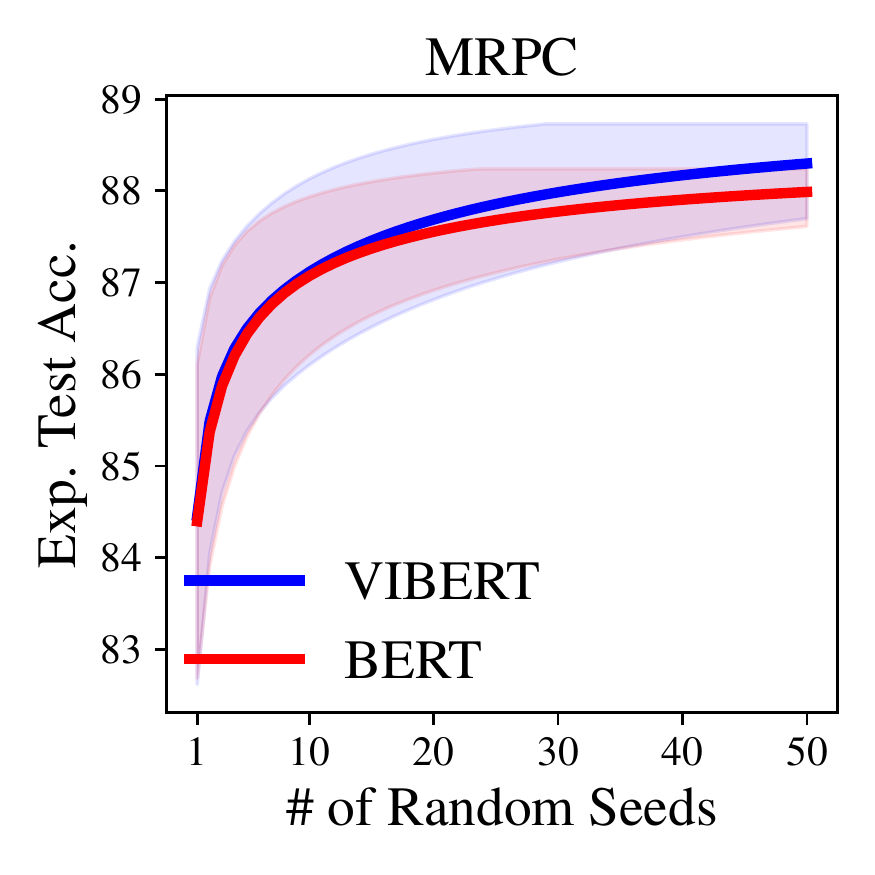}
\end{minipage}\vspace{1em}
\begin{minipage}[t]{.3\linewidth}
    \centering
    \includegraphics[width=\linewidth]{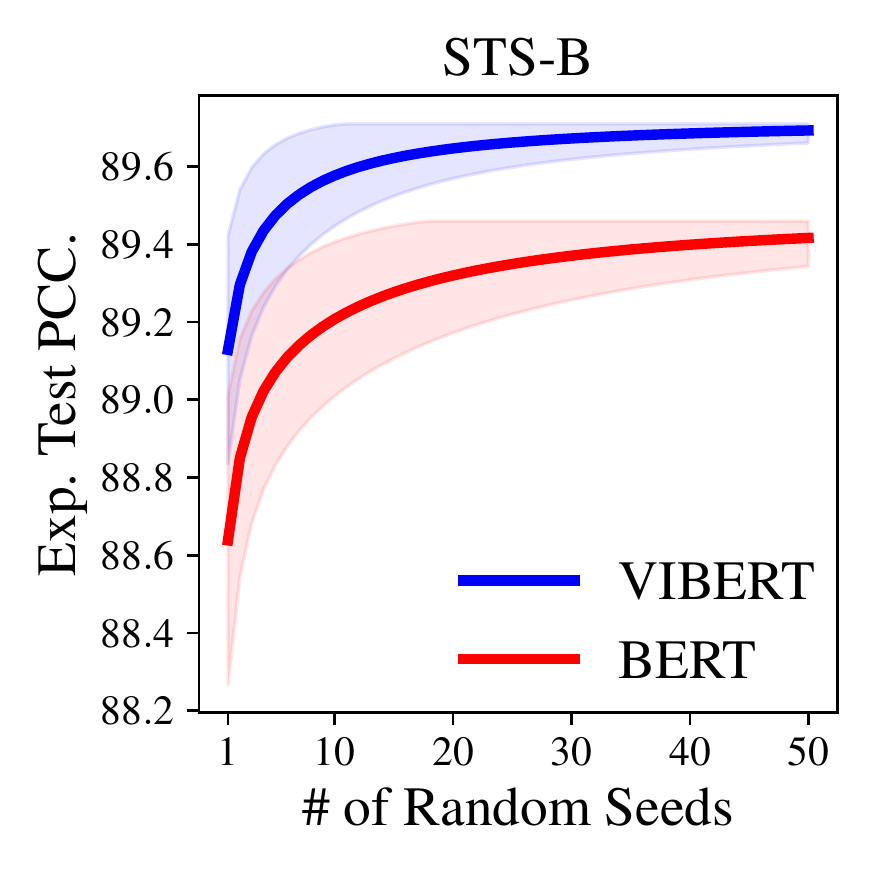}
\end{minipage}\vspace{1em}
\begin{minipage}[t]{.3\linewidth}
    \centering
    \includegraphics[width=\linewidth]{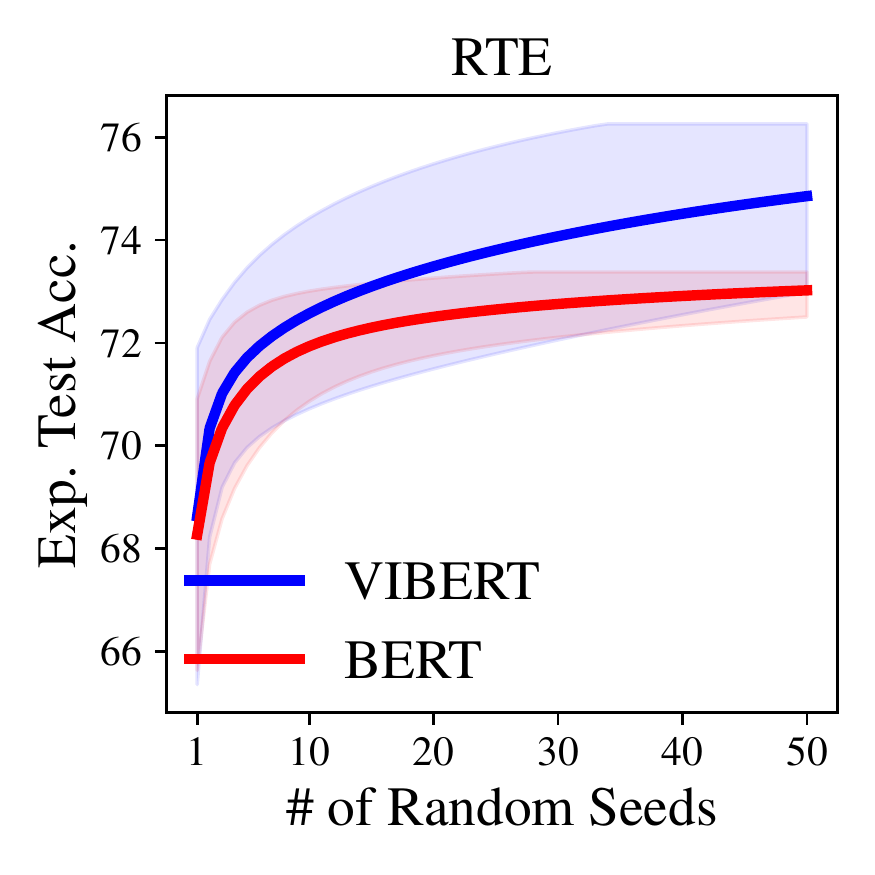}
\end{minipage}
\vspace{-3em}
\caption{Expected test performance (solid lines) with standard deviation (shaded region) over the number of random seeds allocated for fine-tuning. Our VIBERT model consistently outperforms BERT. We report the accuracy for RTE and MRPC and the Pearson correlation coefficient for STS-B.} \vspace{-1em}
\label{fig:random_trials}
\end{figure}

 \paragraph{Impact of Random Seeds:}Following~\cite{dodge2020fine}, we examine the choice of random seed and evaluate the performance of VIBERT and BERT by fine-tuning them across 50 random seeds on GLUE. To comply with the limited access to the GLUE benchmark online system, we split the original validation sets into half and consider one half as the validation set and use the other half as the test set. We first perform model selection on the validation set to fix the hyper-parameters and then fine-tune the selected models for 50 different seeds. Figure~\ref{fig:random_trials} shows the expected test performance~\citep{dodge2019show} as the function of random trials. The results demonstrate that our VIBERT model consistently obtains better performance than BERT on all datasets. As anticipated, the expected test performance monotonically increases with more random trials~\citep{dodge2020fine} till it reaches a plateau, such as after 30 trials on STS-B. 

\subsection{Varying-resource Results} \label{sec:low-resource}
To analyze the performance of our method as a function of dataset size, we use four large-resource NLI and sentiment analysis datasets, namely SNLI, MNLI, IMDB, and YELP to be able to subsample the training data with varying sizes. Table~\ref{tab:sampled_results_test} shows the obtained results. VIBERT consistently outperforms all the baselines on low-resource scenarios, but the advantages are reduced or eliminated as we approach a medium-resource scenario. Also, the improvements are generally larger when the datasets are smaller, showing that our method successfully addresses low-resource scenarios.

\begin{table}[H] \vspace{-0.5em}
    \caption{Test accuracies in the low-resource setting on text classification and NLI datasets under varying sizes of training data (200, 500, 800, 1000, 3000, and 6000 samples). We report the average and standard deviation in parentheses across three runs. We show the highest average result in each setting in bold. $\bm{\Delta}$ shows the absolute difference between the results of VIBERT with BERT.} \vspace{-1em}
    \centering 
    \resizebox{1.0\textwidth}{!}{
    \vspace{-1em}
    \begin{tabular}{l|l|llllll}
    \toprule 
\textbf{Data}     &  \textbf{Model} & $\bm{200}$ & $\bm{500}$& $\bm{800}$ & $\bm{1000}$ & $\bm{3000}$ & $\bm{6000}$ \\
    \toprule      
    \multirow{6}{*}{SNLI} & BERT & 58.70 (1.3) & 68.12 (1.5) & 73.29 (0.9) & 74.69 (1.1) & 79.57 (0.4) & 80.85 (0.4)\\ 
    &~~+Dropout & 58.95 (0.4)  & 69.33 (1.1)&73.22 (1.2) & 74.20 (0.5) & 79.48 (0.7)  & \textbf{81.71} (0.6) \\
    &~~+Mixout & 58.52 (1.3) & 68.26 (1.7) & 72.81 (1.0) & 74.09 (0.5) & 78.7 (0.3) & 80.61 (0.5) \\ 
    &~~+WD & 59.23 (1.5) & 68.54 (1.9) & 73.72 (1.0) & 74.78 (0.8) & \textbf{79.83} (0.5) & 81.32 (0.5) \\ 
    \cmidrule(r){2-8}
    &VIBERT &\textbf{61.42} (1.3) &\textbf{70.75} (0.6) & \textbf{74.71} (0.5) & \textbf{75.84} (0.1) & 79.56 (0.3) & 81.29 (0.4)\\
    & $\bm{\Delta}$ &  +2.72 &  +2.63 &  +1.42 &  +1.15 & -0.01&  +0.44\\
    \toprule 
     \multirow{6}{*}{MNLI} & BERT & 49.93 (1.4) & 59.76 (2.0)  & 63.63 (1.6) & 65.21 (1.4) & 70.67 (0.7) & 73.11 (0.9)\\
     &~~+Dropout & 50.74 (2.1) & 59.58 (2.1)  & 62.82 (0.8) & 65.71 (1.4) & 71.11 (0.8)  & 72.88 (1.1) \\ 
     &~~+Mixout & 50.05 (1.8) & 58.69 (2.8) & 63.31 (1.7) & 64.58 (1.5) & 70.60 (0.8) & 72.56 (0.7) \\
    &~~+WD & 49.92 (1.4) & 60.36 (2.0) & 64.41 (1.5) & 65.3 (1.0) & 71.47 (0.8) & 72.94 (0.7) \\ 
     \cmidrule(r){2-8}
     &VIBERT & \textbf{53.58} (0.9)  & \textbf{63.04} (1.1) & \textbf{64.87} (0.6) & \textbf{66.41} (1.2) & \textbf{71.86} (0.9)& \textbf{74.22} (0.3) \\
     & $\bm{\Delta}$ & +3.65 & +3.28 & +1.24 & +1.2 & +1.19 & +1.11\\ 
    \toprule 
    \multirow{6}{*}{IMDB}  &BERT& 78.96 (1.9) & 83.68 (0.2)  &  84.04 (0.9) & 84.80 (0.0) & 86.17 (0.2) & 86.98 (0.4) \\
    &~~+Dropout & 81.19 (1.6) & 83.30 (0.2)  & 84.52 (0.3) & 85.01 (0.3) & 86.20 (0.2) & \textbf{87.31} (0.2) \\ 
    &~~+Mixout & 79.17 (4.2) & 83.55 (0.3) & 84.37 (0.3) & 84.50 (0.1) & 86.15 (0.1) & 86.97 (0.1) \\ 
    &~~+WD & 79.78 (2.2) & 83.95 (0.2) & 84.29 (0.6) & 84.97 (0.2) & 86.13 (0.3) & 87.2 (0.1) \\ 
     \cmidrule(r){2-8}
    &VIBERT&\textbf{83.05} (0.3)  & \textbf{84.46} (0.4)  & \textbf{84.83} (0.4) & \textbf{85.03} (0.4) & \textbf{86.27} (0.4) & 87.15 (0.3)\\ 
    &$\bm{\Delta}$ & +4.09 &  +0.78 &  +0.79 &  +0.23 & +0.1 & +0.17\\
    \toprule 
    \multirow{6}{*}{YELP} &BERT & 41.60 (0.9) & 44.12 (1.4)  & 45.67 (1.6) & 46.77 (0.5) & 50.14 (0.7) &  51.86 (0.4) \\ 
    &~~+Dropout & 41.30 (0.3) & 44.37 (0.6) & 46.49 (0.8) & 46.21 (1.5) & \textbf{51.09} (0.2) & \textbf{52.39} (0.5) \\
    &~~+Mixout & 41.52 (0.9) & 43.60 (1.1) & 45.65 (1.9) & 46.98 (1.1) & 50.68 (0.5) & 51.51 (0.3) \\ 
    &~~+WD & 41.66 (0.6) & 44.43 (1.2) & 46.26 (1.4) & 47.37 (0.6) & 50.7 (0.5) & 51.9 (0.6)\\ 
    \cmidrule(r){2-8} 
    &VIBERT & \textbf{42.30} (0.2)  & \textbf{46.65} (0.5) & \textbf{46.60} (0.1) & \textbf{48.03} (0.6) & 50.37 (0.4) & 51.34 (0.4)\\
   &$\bm{\Delta}$ &+0.7 &  +2.53 &  +0.93 & +1.26 &  +0.23 & -0.52 \\
   \bottomrule
    \end{tabular}}
    \label{tab:sampled_results_test} \vspace{-0.5em} 
    \end{table}
    
\subsection{Out-of-domain Generalization} \label{sec:lr_transfer}
Besides improving fine-tuning on low-resource data by removing irrelevant features, we expect VIB to improve on out-of-domain data because it removes redundant features. 
In particular, annotation artifacts create shortcut features, which are superficial cues correlated with a label~\citep{gururangan2018annotation, poliak2018hypothesis} that do not generalize well to out-of-domain datasets~\citep{belinkov-etal-2019-dont1}.
Since solving the real underlying task can be done without these superficial shortcuts, they must be redundant with the deep semantic features that are truly needed.  We hypothesize that many more superficial shortcut features are needed to reach the same level of performance as a few deep semantic features.  If so, then VIB should prefer to keep the concise deep features and remove the abundant superficial features, 
thus encouraging the classifier to rely on the deep semantic features, and therefore resulting in better generalization to out-of-domain data.
To evaluate out-of-domain generalization, we take  NLI models trained on medium-sized 6K subsampled SNLI and MNLI in Section~\ref{sec:low-resource} and evaluate their generalization on several NLI datasets.

\paragraph{Datasets:} We consider a total of 15 different NLI datasets used in~\citet{karimi2020bias}, including SICK~\citep{MARELLI14.363}, ADD1~\citep{pavlick2016most}, JOCI~\citep{zhang2017ordinal}, MPE~\citep{lai2017natural},  MNLI, SNLI, SciTail~\citep{khot2018scitail}, and three datasets from \citet{white2017inference} namely DPR~\citep{rahman2012resolving}, FN+~\citep{pavlick2015framenet+}, SPR~\citep{reisinger2015semantic}, and Quora Question Pairs (QQP) interpreted as an NLI task as by \citet{gong2017natural}.  We use the same split used in~\citet{wang2017bilateral}. We also consider SNLI hard and MNLI(-M) Hard sets~\citep{gururangan2018annotation}, a subset of SNLI/MNLI(-M) where a hypothesis-only model cannot correctly predict the labels and the known biases are avoided.
Since the target datasets have different label spaces, during the evaluation, we map predictions to each target dataset's space (Appendix~\ref{app:transfer}). Following prior work~\citep{belinkov-etal-2019-dont1, karimi2020bias},  we select hyper-parameters based on the development set of each target dataset and report the results on the test set. 

\noindent\paragraph{Results:} Table~\ref{tab:transfer_results} shows the results of VIBERT and BERT. We additionally include WD, the baseline that performed the best on average on SNLI and MNLI in Table~\ref{tab:sampled_results_test}. On models trained on SNLI, VIBERT improves the transfer on 13 out of 15 datasets, obtaining a substantial average improvement of 5.51 points. The amount of improvement on different datasets varies, with the largest improvement on SPR and SciTail with +15.5, and +12.5 points respectively, while WD on average obtains only 0.99 points improvement. On models trained on MNLI, VIBERT improves the transfer on 13 datasets, obtaining an average improvement of 3.83 points. The improvement varies across the datasets, with the largest on ADD1 and JOCI with 16.8 and 8.3 points respectively, substantially surpassing WD. Interestingly, VIBERT improves the results on the SNLI and MNLI(-M) hard sets, resulting in models that are more robust to known biases. These results support our claim that VIBERT motivates learning more general features, rather than redundant superficial features, leading to an improved generalization to datasets without these superficial biases. In the next section, we analyze this phenomenon more.

\begin{table}[H] 
\vspace{-0.5em}
 \caption{Test accuracy of models transferring to new target datasets. All models are trained on SNLI or MNLI and tested on the target datasets. \bm{$\Delta$} are absolute differences with BERT.}
    \centering \vspace{-0.8em}
    \resizebox{1.0\textwidth}{!}{
    \begin{tabular}{llllllllllllll}
        \toprule 
        & \multicolumn{5}{c}{\textbf{SNLI}} & \multicolumn{5}{c}{\textbf{MNLI}} \\ 
         \cmidrule(r){2-6} \cmidrule(l){7-11} 
        {\vspace{-0.75em} \textbf{Data}} &  \multicolumn{2}{c}{} & \multicolumn{2}{c}{}\\ 
                 & \rotatebox{0}{\textbf{\footnotesize{BERT}}} & \rotatebox{0}{\textbf{\footnotesize{VIBERT}}} & \bm{$\Delta$} & \rotatebox{0}{\textbf{\footnotesize{WD}}}& \bm{$\Delta$} &\textbf{\footnotesize{BERT}} & \textbf{\footnotesize{VIBERT}} & \bm{$\Delta$}& \textbf{\footnotesize{WD}}& \bm{$\Delta$}\\  
        \toprule  
        SICK & 48.47& 54.68& +6.2 & 48.37 & -0.1& 59.16 & 69.17&+10.0 & 63.87 &+4.7\\
        ADD1 & 78.81& 84.75&+5.9  & 80.62& +1.8& 66.15 &82.95 & +16.8 &67.18 &+1.0 \\
        DPR & 50.78 & 50.14&-0.6   &50.41 &-0.4 & 49.95 &49.95 & 0.0 &49.95 & 0.\\
        SPR& 50.21  & 65.68&+15.5  &51.90 &+1.7 & 59.16 & 65.61& +6.5 & 57.21& -1.9 \\
        FN+ & 50.78 & 53.44&+2.7   &50.58  &-0.2 & 46.28 &49.94 &+3.7 &46.34 & +0.1 \\
        JOCI & 42.03 &50.66 &+8.6   &43.91  &+1.9 & 45.60 &53.94 &+8.3 & 46.49& +0.9 \\
        MPE &  58.30 & 58.10 &-0.2  &58.10  &-0.2 & 55.10 &50.30 & -4.8 &58.2 & +3.1 \\
        SCITAIL&62.32 &74.84 &+12.5 &65.10  &+2.8 &72.58 &75.68 & +3.1 &75.73 & +3.2  \\
        QQP & 65.19 & 70.67&+5.5    &65.90  &+0.7 &67.88 &70.50 &+2.6 & 68.75& +0.9 \\
        \midrule 
        SNLI Hard & 65.72&68.35 &+2.6  &66.82  &+1.1 &56.98 &60.29 & +3.3 & 57.8 & +0.8\\
        MNLI Hard & 46.31  & 53.17 & +6.9 &47.42   & +1.1& 59.74 &61.19& +1.4& 60.08& +0.3\\ 
        MNLI-M Hard &  46.12 & 52.38 & +6.3 &46.82 & +0.7 & 60.55  &61.03 & +0.5& 59.77 &-0.8  \\  
        \midrule 
        SNLI & 80.54 &81.81 &+1.3  &81.26 &+0.7 & 64.32 &67.87 &+3.6 & 65.44 & +1.1 \\ 
        MNLI-M & 60.51 &64.88 &+4.4 &62.11  &+1.6 & 72.42 &73.06 &+0.6 &72.76& +0.3\\
        MNLI &61.79 &66.76 &+5.0  & 63.42 &+1.6 & 72.73 &74.67 & +1.9& 72.89& +0.2\\
        \midrule 
        Average &--- & ---& +5.51 & --- & +0.99 & --- & --- & +3.83 & --- & +0.93 \\ 
    \bottomrule 
    \end{tabular}}
\vspace{-1.2em}
    \label{tab:transfer_results}
\end{table}

\section{Analysis}
\paragraph{Analysis of the Removed Features}\citet{elazar2018adversarial} propose a challenging framework to evaluate if debiasing methods have succeeded in removing biases from the sentence representation.  After debiasing, the trained encoder is frozen and the classifier is retrained to try to extract the biases.  If the classifier reaches high accuracy given only bias features, then the encoder's representation has not been successfully debiased.
We follow the framework of~\citet{elazar2018adversarial} to analyze whether known biases in NLI data have been removed in the trained sentence representations.  In particular, following~\citet{belinkov2019adversarial}, we train a classifier which only sees the representation of the hypothesis sentence and see if it can predict the class of the sentence pair, which is an established criterion to measure known biases in NLI datasets~\citep{gururangan2018annotation}.  Thus, we freeze the trained encoders from our model and the BERT baseline and retrain a hypothesis-only classifier on hypotheses from the SNLI and MNLI datasets.\footnote{Note that with VIBERT, the frozen encoder $p_{\theta}(z|x)$ outputs a distribution, and the hypothesis-only classifier is trained on samples from this distribution.}  For reference, we compare to a hypothesis-only model with a BERT encoder trained end-to-end.  Table~\ref{tab:bias_results} shows the results.
With the baseline (BERT), the retrained classifier is not able to recapture all the biases (H-only), but it captures much more than with our method (VIBERT).  VIBERT is so successful at reducing biases that performance of the hypothesis-only classifier is close to chance (33\%).

\begin{table}[H]\vspace{-0.7em}
    \caption{Hypothesis-only accuracy when freezing the encoder from models trained on SNLI/MNLI in Table~\ref{tab:sampled_results_test} and retraining a hypothesis-only classifier (BERT, VIBERT), and baseline results when the encoder is not frozen (H-only). Lower results show more successful debiasing.
    \vspace{-0.5ex}}
    \centering
    \begin{tabular}{lllllll}
    \toprule 
     \multirow{2}{*}{\textbf{Model}} & \multicolumn{3}{c}{\textbf{SNLI}} & \multicolumn{3}{c}{\textbf{MNLI}} \\ 
         \cmidrule(r){2-4} \cmidrule(l){5-7} 
      &   \textbf{Train}  & \textbf{Dev} & \textbf{Test} &  \textbf{Train}  & \textbf{Dev} & \textbf{Test} \\
     \toprule
    H-only & 81.3 & 61.89 & 62.17 & 87.15 & 53.46 & 53.63\\ 
    \midrule 
    BERT &   66.40 & 53.73 &  53.17  &  58.5 &  44.68 & 44.03\\
    VIBERT &  38.20 &  \textbf{36.65} &   \textbf{37.10} & 42.03 &  \textbf{36.43} & \textbf{35.75} \\
    \bottomrule
    \end{tabular}
\vspace{-0.75em}
    \label{tab:bias_results}
\end{table} 

\begin{figure}[H]
\centering 
\begin{minipage}[t]{.33\linewidth}
    \centering
    \includegraphics[width=\linewidth]{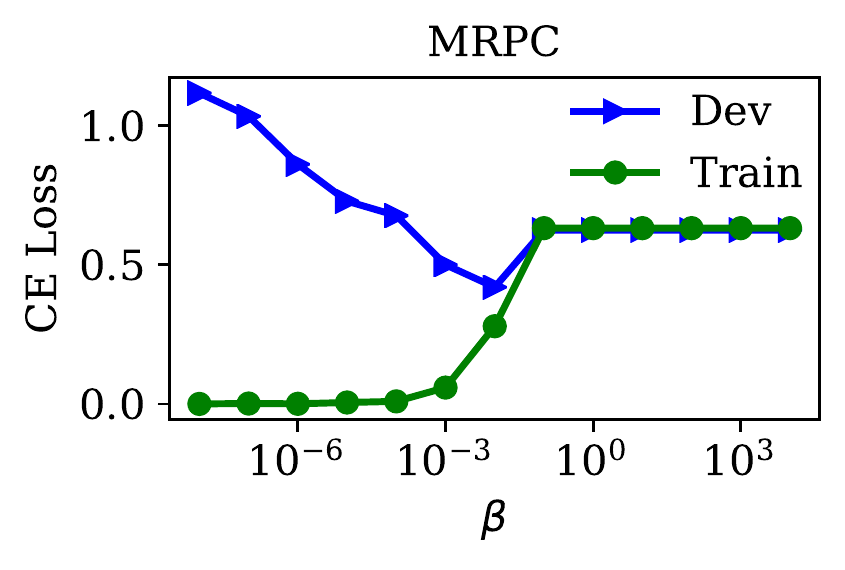}
\end{minipage}\hfill
\begin{minipage}[t]{.33\linewidth}
    \centering
    \includegraphics[width=\linewidth]{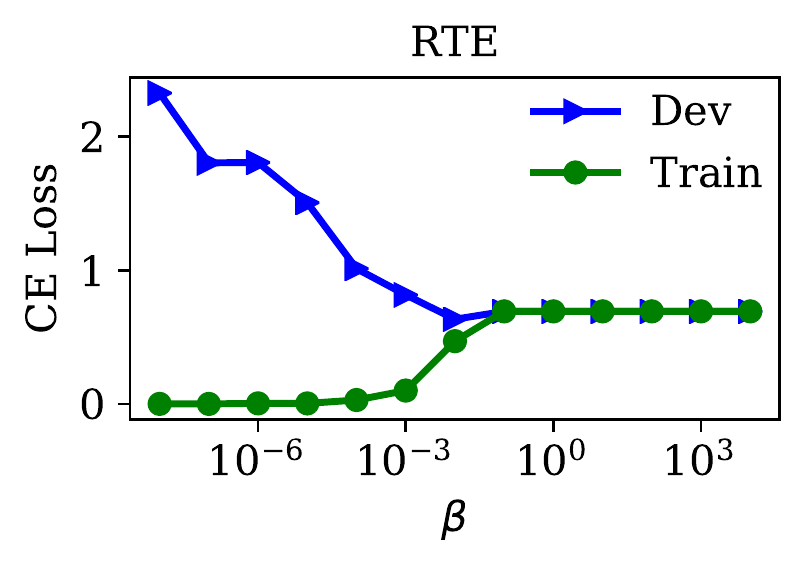}
\end{minipage}\hfill
\begin{minipage}[t]{.33\linewidth}
    \centering
    \includegraphics[width=\linewidth]{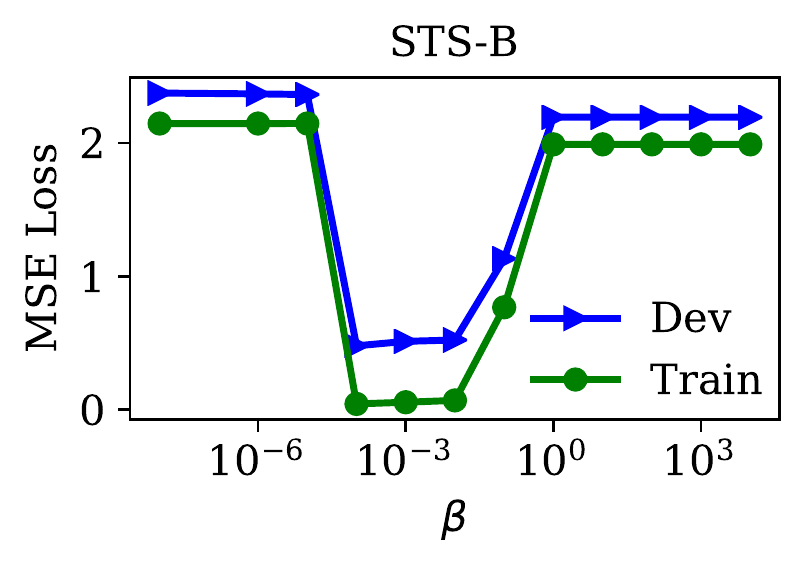}
\end{minipage}\vspace{-1em}
\caption{Validation and training losses of VIBERT for varying $\beta$ and a fixed bottleneck size on GLUE.} 
\vspace{-1em}
\label{fig:varying_beta}
\end{figure}

\paragraph{Impact of VIB on Overfitting}
To analyze the effect of VIB  on reducing overfitting, we analyze the effect of the $\beta$ parameter on training and validation error since $\beta$ controls the trade-off between removing information from the sentence embedding (high $\beta$) and keeping information that is predictive of the output (low $\beta$). We fix the bottleneck size ($K$) based on the models selected in Section~\ref{sec:glue_results}, and we train VIBERT on the GLUE benchmark for varying values of $\beta$ and plot the validation and training loss in Figure~\ref{fig:varying_beta}. 

For small values of $\beta$, where VIB has little effect, the validation loss is substantially higher than the training loss, indicating overfitting. This is because the network learns to be more deterministic ($\Sigma \approx 0$), thereby retaining too much irrelevant information. As we increase $\beta$, where VIB has an effect, we observe better generalization performance with less overfitting.  As $\beta$ becomes too large, both the training and validation losses shoot up because the amount of preserved information is insufficient to differentiate between the classes. This pattern is observable in the MRPC and RTE datasets, with a similar pattern in the STS-B dataset. 

\paragraph{Efficiency Evaluation} 
Table~\ref{tab:performance} presents the efficiency evaluation in terms of memory, number of parameters, and time for all the methods measured on RTE. Our approach has several attractive properties. First, while our method is slightly larger in terms of parameters compared to the other standard regularization approaches due to an additional MLP layer (Figure~\ref{fig:IB}), the difference is still marginal, and for BERT$_\text{Base}$ model with 109.48M trainable parameters, that is less than 1.22\% more parameters. Second, our approach presents a much better memory usage with low-overhead, close to Dropout, while WD and especially Mixout cause substantial memory overhead. In dealing with large-scale transformer models like BERT, efficient memory usage is of paramount importance. Third, in terms of training time, our method is similar to Dropout and much faster than the other two baselines. Relative to BERT, VIBERT increases the training time by 3.77\%, while WD and Mixout cause the substantial training overhead of 9.11\% and 14.44\%. Note that our method and other baselines require hyper-parameter tuning. 

\begin{table}[t!]
    \caption{Performance evaluation for all methods. $\bm{\Delta\%}$ are relative differences with BERT.}\vspace{-1em}
\centering 
    \begin{tabular}{l|lr|lr|lr}
    \toprule 
    \textbf{Model} & \textbf{Memory} & \bm{$\Delta$\%} & \#\textbf{Parameters} &\bm{$\Delta$\%} & \textbf{Time} &\bm{$\Delta$\%}\\
    \toprule 
    BERT      & 290.91 GB  &---       & 109.48 M & --- & 4.50 min & --- \\
     ~+Mixout  & 407.65 GB  & 40.13 \% & 109.48 M & 0\% & 5.15 min & 14.44\% \\ 
     ~+WD      & 331.78 GB  & 14.05\%  & 109.48 M & 0\% & 4.91 min & 9.11\%\\ 
     ~+Dropout & 290.91 GB  & 0\%      & 109.48 M & 0\% & 4.68 min & 4\% \\ 
     \midrule 
     VIBERT & 292.57 GB  & 0.57 \% & 110.83 M & 1.22\% &  4.67 min& 3.77\%\\
    \bottomrule
    \end{tabular}
\label{tab:performance}
\end{table}

\paragraph{Ablation Study}
As an ablation, Table~\ref{tab:glue_results_ablation} shows results for our model without the compression loss (VIBERT ($\beta=0$)), in which case there is no incentive to introduce noise, and the VIB layer reduces to deterministic dimensionality reduction with an MLP. We optimize the dimensionality of the MLP layer ($K$) as a hyper-parameter for both methods. This ablation does reduce performance on all considered datasets, demonstrating the added benefit of the compression loss of VIBERT.

\begin{table}[H]
 \caption{Average ablation results over 3 runs with std in parentheses on GLUE. BERT and VIBERT's results are from Table~\ref{tab:glue_results}.} \vspace{-1em}
\centering
      \begin{center}
    \begin{tabular}{lllllll} 
        \toprule 
         & \multicolumn{2}{c}{\textbf{MRPC}} & \multicolumn{2}{c}{\textbf{STS-B}} & \multicolumn{1}{c}{\textbf{RTE}}\\ 
         \cmidrule(r){2-3} \cmidrule(l){4-5}  \cmidrule(l){6-6}
        {\vspace{-0.75em} \textbf{Model}} &  \multicolumn{2}{c}{} & \multicolumn{2}{c}{} &\\ 
                 & \textbf{Accuracy} & \textbf{F1}  &\textbf{Pearson} & \textbf{Spearman} & \textbf{Accuracy}\\  
        \toprule 
    BERT & 87.80 (0.5) & 83.20 (0.6) & 84.93 (0.1) & 83.53 (0.0) & 67.93 (1.5) \\ 
    \midrule 
    VIBERT ($\beta$=0) &88.57 (0.6) &84.27 (0.7) & 87.10 (0.4) & 86.00 (0.5) & 69.63 (1.3) \\ 
   VIBERT &\textbf{89.23} (0.1) & \textbf{85.23} (0.2) & \textbf{87.63} (0.3) & \textbf{86.50} (0.4) & \textbf{70.53} (0.5) \\ 
    \bottomrule
    \end{tabular}
    \end{center}
    \label{tab:glue_results_ablation} \vspace{-1.5em}
\end{table}

\section{Related Work}

\vspace{-0.5ex}

\paragraph{Low-resource Setting} Recently, developing methods for low-resource NLP has gained attention~\citep{emnlp-2019-deep}. Prior work has investigated improving on low-resource datasets by injecting large unlabeled in-domain data and pretraining a unigram document model using a variational autoencoder and use its internal representations as features for downstream tasks~\citep{gururangan2019variational}. Other approaches propose injecting a million-scale previously collected phrasal paraphrase relations~\citep{arase2019transfer} and data augmentation for translation task~\citep{fadaee2017data}.  Due to relying on the additional source and in-domain corpus, such techniques are not directly comparable to our model.

\vspace{-0.5ex}

\paragraph{Information Bottleneck} IB has recently been adopted in NLP in applications such as parsing~\citep{li2019specializing}, and summarization~\citep{west2019bottlesum}.~\citet{voita2019bottom} use the mutual information to study how token representations evolve across layers of a Transformer model~\citep{vaswani2017attention}. This paper – to the best of our knowledge – is the first attempt to study VIB as a regularization technique to improve the fine-tuning of large-scale language models on low-resource scenarios.

\vspace{-0.5ex}

\paragraph{Regularization Techniques for Fine-tuning Language models}
In addition to references given throughout,~\citet{phang2018sentence} proposed to perform an extra data-rich intermediate supervised task pretraining followed by fine-tuning on the target task. They showed that their method leads to improved fine-tuning performance on the GLUE benchmark. However, their method requires pretraining with a large intermediate task. In contrast, our goal is to use only the provided low-resource target datasets.

\section{Conclusion and Future Directions} 
We propose VIBERT, an effective model to reduce overfitting when fine-tuning large-scale pretrained language models on low-resource datasets. By leveraging a VIB objective, VIBERT finds the simplest sentence embedding, predictive of the target labels, while removing task-irrelevant and redundant information. Our approach is model agnostic, simple to implement, and highly effective. Extensive experiments and analyses show that our method substantially improves transfer performance in low-resource scenarios. We demonstrate our obtained sentence embeddings are robust to biases and our model results in a substantially better generalization to out-of-domain NLI datasets. Future work includes exploring incorporating VIB on multiple layers of pretrained language models and using it to jointly learn relevant features and relevant layers.

\section*{Acknowledgements}
We would like to thank Maksym Andriushchenko for his helpful comments. Rabeeh Karimi was supported by the Swiss National
Science Foundation under the project Learning Representations of Abstraction for Opinion Summarisation (LAOS), grant number “FNS-30216”. Yonatan Belinkov was supported by the ISRAEL SCIENCE FOUNDATION (grant No. 448/20).

\bibliography{iclr2021_conference}

\begin{thebibliography}{57}
\providecommand{\natexlab}[1]{#1}
\providecommand{\url}[1]{\texttt{#1}}
\expandafter\ifx\csname urlstyle\endcsname\relax
  \providecommand{\doi}[1]{doi: #1}\else
  \providecommand{\doi}{doi: \begingroup \urlstyle{rm}\Url}\fi

\bibitem[Alemi et~al.(2017)Alemi, Fischer, Dillon, and Murphy]{alemi2016deep}
Alex Alemi, Ian Fischer, Josh Dillon, and Kevin Murphy.
\newblock Deep variational information bottleneck.
\newblock In \emph{ICLR}, 2017.

\bibitem[Arase \& Tsujii(2019)Arase and Tsujii]{arase2019transfer}
Yuki Arase and Jun’ichi Tsujii.
\newblock Transfer fine-tuning: A bert case study.
\newblock In \emph{EMNLP}, 2019.

\bibitem[Belinkov et~al.(2019{\natexlab{a}})Belinkov, Poliak, Shieber,
  Van~Durme, and Rush]{belinkov-etal-2019-dont1}
Yonatan Belinkov, Adam Poliak, Stuart Shieber, Benjamin Van~Durme, and
  Alexander Rush.
\newblock Don{'}t take the premise for granted: Mitigating artifacts in natural
  language inference.
\newblock In \emph{ACL}, 2019{\natexlab{a}}.

\bibitem[Belinkov et~al.(2019{\natexlab{b}})Belinkov, Poliak, Shieber,
  Van~Durme, and Rush]{belinkov2019adversarial}
Yonatan Belinkov, Adam Poliak, Stuart~M Shieber, Benjamin Van~Durme, and
  Alexander~M Rush.
\newblock On adversarial removal of hypothesis-only bias in natural language
  inference.
\newblock In \emph{SEM}, 2019{\natexlab{b}}.

\bibitem[Bowman et~al.(2016)Bowman, Vilnis, Vinyals, Dai, Jozefowicz, and
  Bengio]{bowman2016generating}
Samuel Bowman, Luke Vilnis, Oriol Vinyals, Andrew Dai, Rafal Jozefowicz, and
  Samy Bengio.
\newblock Generating sentences from a continuous space.
\newblock In \emph{CoNLL}, 2016.

\bibitem[Bowman et~al.(2015)Bowman, Angeli, Potts, and
  Manning]{bowman2015large}
Samuel~R. Bowman, Gabor Angeli, Christopher Potts, and Christopher~D. Manning.
\newblock A large annotated corpus for learning natural language inference.
\newblock In \emph{EMNLP}, 2015.

\bibitem[Cer et~al.(2017)Cer, Diab, Agirre, Lopez-Gazpio, and
  Specia]{cer2017semeval}
Daniel Cer, Mona Diab, Eneko Agirre, I{\~n}igo Lopez-Gazpio, and Lucia Specia.
\newblock Semeval-2017 task 1: Semantic textual similarity multilingual and
  crosslingual focused evaluation.
\newblock In \emph{{S}em{E}val}, 2017.

\bibitem[Chelba \& Acero(2004)Chelba and Acero]{chelba2004adaptation}
Ciprian Chelba and Alex Acero.
\newblock Adaptation of maximum entropy capitalizer: Little data can help a
  lot.
\newblock In \emph{EMNLP}, 2004.

\bibitem[Cherry et~al.(2019)Cherry, Durrett, Foster, Haffari, Khadivi, Peng,
  Ren, and Swayamdipta]{emnlp-2019-deep}
Colin Cherry, Greg Durrett, George Foster, Reza Haffari, Shahram Khadivi,
  Nanyun Peng, Xiang Ren, and Swabha Swayamdipta (eds.).
\newblock \emph{DeepLo}, 2019.
\newblock URL \url{https://www.aclweb.org/anthology/D19-6100}.

\bibitem[Dagan et~al.(2006)Dagan, Glickman, and Magnini]{dagan2006pascal}
Ido Dagan, Oren Glickman, and Bernardo Magnini.
\newblock The pascal recognising textual entailment challenge.
\newblock In \emph{MLCW}, 2006.

\bibitem[Daum{\'e}~III(2007)]{daume2007frustratingly}
Hal Daum{\'e}~III.
\newblock Frustratingly easy domain adaptation.
\newblock In \emph{ACL}, 2007.

\bibitem[Devlin et~al.(2019)Devlin, Chang, Lee, and Toutanova]{devlin2019bert}
Jacob Devlin, Ming-Wei Chang, Kenton Lee, and Kristina Toutanova.
\newblock Bert: Pre-training of deep bidirectional transformers for language
  understanding.
\newblock In \emph{NAACL}, 2019.

\bibitem[Dodge et~al.(2019)Dodge, Gururangan, Card, Schwartz, and
  Smith]{dodge2019show}
Jesse Dodge, Suchin Gururangan, Dallas Card, Roy Schwartz, and Noah~A Smith.
\newblock Show your work: Improved reporting of experimental results.
\newblock In \emph{EMNLP-IJCNLP}, 2019.

\bibitem[Dodge et~al.(2020)Dodge, Ilharco, Schwartz, Farhadi, Hajishirzi, and
  Smith]{dodge2020fine}
Jesse Dodge, Gabriel Ilharco, Roy Schwartz, Ali Farhadi, Hannaneh Hajishirzi,
  and Noah Smith.
\newblock Fine-tuning pretrained language models: Weight initializations, data
  orders, and early stopping.
\newblock \emph{arXiv:2002.06305}, 2020.

\bibitem[Dolan \& Brockett(2005)Dolan and Brockett]{dolan2005automatically}
William~B Dolan and Chris Brockett.
\newblock Automatically constructing a corpus of sentential paraphrases.
\newblock In \emph{IWP}, 2005.

\bibitem[Elazar \& Goldberg(2018)Elazar and Goldberg]{elazar2018adversarial}
Yanai Elazar and Yoav Goldberg.
\newblock Adversarial removal of demographic attributes from text data.
\newblock In \emph{EMNLP}, 2018.

\bibitem[Fadaee et~al.(2017)Fadaee, Bisazza, and Monz]{fadaee2017data}
Marzieh Fadaee, Arianna Bisazza, and Christof Monz.
\newblock Data augmentation for low-resource neural machine translation.
\newblock In \emph{ACL}, 2017.

\bibitem[Gong et~al.(2017)Gong, Luo, and Zhang]{gong2017natural}
Yichen Gong, Heng Luo, and Jian Zhang.
\newblock Natural language inference over interaction space.
\newblock In \emph{ICLR}, 2017.

\bibitem[Gururangan et~al.(2018)Gururangan, Swayamdipta, Levy, Schwartz,
  Bowman, and Smith]{gururangan2018annotation}
Suchin Gururangan, Swabha Swayamdipta, Omer Levy, Roy Schwartz, Samuel Bowman,
  and Noah~A. Smith.
\newblock Annotation artifacts in natural language inference data.
\newblock In \emph{NAACL}, 2018.

\bibitem[Gururangan et~al.(2019)Gururangan, Dang, Card, and
  Smith]{gururangan2019variational}
Suchin Gururangan, Tam Dang, Dallas Card, and Noah~A Smith.
\newblock Variational pretraining for semi-supervised text classification.
\newblock In \emph{ACL}, 2019.

\bibitem[Harremo{\"e}s \& Tishby(2007)Harremo{\"e}s and
  Tishby]{harremoes2007information}
Peter Harremo{\"e}s and Naftali Tishby.
\newblock The information bottleneck revisited or how to choose a good
  distortion measure.
\newblock In \emph{ISIT}, 2007.

\bibitem[Igl et~al.(2019)Igl, Ciosek, Li, Tschiatschek, Zhang, Devlin, and
  Hofmann]{igl2019generalization}
Maximilian Igl, Kamil Ciosek, Yingzhen Li, Sebastian Tschiatschek, Cheng Zhang,
  Sam Devlin, and Katja Hofmann.
\newblock Generalization in reinforcement learning with selective noise
  injection and information bottleneck.
\newblock In \emph{NeurIPS}, 2019.

\bibitem[Kaushik \& Lipton(2018)Kaushik and Lipton]{kaushik2018much}
Divyansh Kaushik and Zachary~C Lipton.
\newblock How much reading does reading comprehension require? a critical
  investigation of popular benchmarks.
\newblock In \emph{EMNLP}, 2018.

\bibitem[Khot et~al.(2018)Khot, Sabharwal, and Clark]{khot2018scitail}
Tushar Khot, Ashish Sabharwal, and Peter Clark.
\newblock Scitail: A textual entailment dataset from science question
  answering.
\newblock In \emph{AAAI}, 2018.

\bibitem[Kingma \& Welling(2013)Kingma and Welling]{kingma2013auto}
Diederik~P Kingma and Max Welling.
\newblock Auto-encoding variational bayes.
\newblock In \emph{ICLR}, 2013.

\bibitem[Krogh \& Hertz(1992)Krogh and Hertz]{krogh1992simple}
Anders Krogh and John~A Hertz.
\newblock A simple weight decay can improve generalization.
\newblock In \emph{NeurIPS}, 1992.

\bibitem[Lai et~al.(2017)Lai, Bisk, and Hockenmaier]{lai2017natural}
Alice Lai, Yonatan Bisk, and Julia Hockenmaier.
\newblock Natural language inference from multiple premises.
\newblock In \emph{IJCNLP}, 2017.

\bibitem[Lee et~al.(2019)Lee, Cho, and Kang]{lee2019mixout}
Cheolhyoung Lee, Kyunghyun Cho, and Wanmo Kang.
\newblock Mixout: Effective regularization to finetune large-scale pretrained
  language models.
\newblock In \emph{ICLR}, 2019.

\bibitem[Li \& Eisner(2019)Li and Eisner]{li2019specializing}
Xiang~Lisa Li and Jason Eisner.
\newblock Specializing word embeddings (for parsing) by information bottleneck.
\newblock In \emph{EMNLP}, 2019.

\bibitem[Liu et~al.(2019)Liu, Ott, Goyal, Du, Joshi, Chen, Levy, Lewis,
  Zettlemoyer, and Stoyanov]{liu2019roberta}
Yinhan Liu, Myle Ott, Naman Goyal, Jingfei Du, Mandar Joshi, Danqi Chen, Omer
  Levy, Mike Lewis, Luke Zettlemoyer, and Veselin Stoyanov.
\newblock Roberta: A robustly optimized bert pretraining approach.
\newblock \emph{arXiv:1907.11692}, 2019.

\bibitem[Maas et~al.(2011)Maas, Daly, Pham, Huang, Ng, and
  Potts]{maas2011learning}
Andrew~L. Maas, Raymond~E. Daly, Peter~T. Pham, Dan Huang, Andrew~Y. Ng, and
  Christopher Potts.
\newblock Learning word vectors for sentiment analysis.
\newblock In \emph{ACL}, 2011.

\bibitem[Mahabadi et~al.(2020)Mahabadi, Belinkov, and
  Henderson]{karimi2020bias}
Karimi~Rabeeh Mahabadi, Yonatan Belinkov, and James Henderson.
\newblock End-to-end bias mitigation by modelling biases in corpora.
\newblock In \emph{ACL}, 2020.

\bibitem[Marelli et~al.(2014)Marelli, Menini, Baroni, Bentivogli, Bernardi, and
  Zamparelli]{MARELLI14.363}
Marco Marelli, Stefano Menini, Marco Baroni, Luisa Bentivogli, Raffaella
  Bernardi, and Roberto Zamparelli.
\newblock A sick cure for the evaluation of compositional distributional
  semantic models.
\newblock In \emph{LREC}, 2014.

\bibitem[Mosbach et~al.(2021)Mosbach, Andriushchenko, and
  Klakow]{mosbach2020stability}
Marius Mosbach, Maksym Andriushchenko, and Dietrich Klakow.
\newblock On the stability of fine-tuning bert: Misconceptions, explanations,
  and strong baselines.
\newblock \emph{ICLR}, 2021.

\bibitem[Pavlick \& Callison-Burch(2016)Pavlick and
  Callison-Burch]{pavlick2016most}
Ellie Pavlick and Chris Callison-Burch.
\newblock Most {``}babies{''} are {``}little{''} and most {``}problems{''} are
  {``}huge{''}: Compositional entailment in adjective-nouns.
\newblock In \emph{ACL}, 2016.

\bibitem[Pavlick et~al.(2015)Pavlick, Wolfe, Rastogi, Callison-Burch, Dredze,
  and Van~Durme]{pavlick2015framenet+}
Ellie Pavlick, Travis Wolfe, Pushpendre Rastogi, Chris Callison-Burch, Mark
  Dredze, and Benjamin Van~Durme.
\newblock Framenet+: Fast paraphrastic tripling of framenet.
\newblock In \emph{ACL}, 2015.

\bibitem[Phang et~al.(2018)Phang, F{\'e}vry, and Bowman]{phang2018sentence}
Jason Phang, Thibault F{\'e}vry, and Samuel~R Bowman.
\newblock Sentence encoders on stilts: Supplementary training on intermediate
  labeled-data tasks.
\newblock \emph{arXiv:1811.01088}, 2018.

\bibitem[Poliak et~al.(2018)Poliak, Naradowsky, Haldar, Rudinger, and
  Van~Durme]{poliak2018hypothesis}
Adam Poliak, Jason Naradowsky, Aparajita Haldar, Rachel Rudinger, and Benjamin
  Van~Durme.
\newblock Hypothesis only baselines in natural language inference.
\newblock In \emph{SEMEVAL}, 2018.

\bibitem[Radford et~al.(2019)Radford, Wu, Child, Luan, Amodei, and
  Sutskever]{radford2019language}
Alec Radford, Jeff Wu, Rewon Child, David Luan, Dario Amodei, and Ilya
  Sutskever.
\newblock Language models are unsupervised multitask learners.
\newblock 2019.

\bibitem[Rahman \& Ng(2012)Rahman and Ng]{rahman2012resolving}
Altaf Rahman and Vincent Ng.
\newblock Resolving complex cases of definite pronouns: the winograd schema
  challenge.
\newblock In \emph{EMNPL}, 2012.

\bibitem[Reisinger et~al.(2015)Reisinger, Rudinger, Ferraro, Harman, Rawlins,
  and Van~Durme]{reisinger2015semantic}
Drew Reisinger, Rachel Rudinger, Francis Ferraro, Craig Harman, Kyle Rawlins,
  and Benjamin Van~Durme.
\newblock Semantic proto-roles.
\newblock In \emph{TACL}, 2015.

\bibitem[Schuster et~al.(2019)Schuster, J~Shah, Jie Serene~Yeo, Filizzola,
  Santus, and Barzilay]{schuster2019towards}
Tal Schuster, Darsh J~Shah, Yun Jie Serene~Yeo, Daniel Filizzola, Enrico
  Santus, and Regina Barzilay.
\newblock Towards debiasing fact verification models.
\newblock In \emph{EMNLP}, 2019.

\bibitem[Shamir et~al.(2010)Shamir, Sabato, and Tishby]{shamir2010learning}
Ohad Shamir, Sivan Sabato, and Naftali Tishby.
\newblock Learning and generalization with the information bottleneck.
\newblock In \emph{TCS}, 2010.

\bibitem[Srivastava et~al.(2014)Srivastava, Hinton, Krizhevsky, Sutskever, and
  Salakhutdinov]{srivastava2014dropout}
Nitish Srivastava, Geoffrey Hinton, Alex Krizhevsky, Ilya Sutskever, and Ruslan
  Salakhutdinov.
\newblock Dropout: a simple way to prevent neural networks from overfitting.
\newblock In \emph{JMLR}, 2014.

\bibitem[Tishby et~al.(1999)Tishby, Pereira, and
  Bialek]{Tishby99theinformation}
Naftali Tishby, Fernando~C. Pereira, and William Bialek.
\newblock The information bottleneck method.
\newblock In \emph{Allerton}, 1999.

\bibitem[Vaswani et~al.(2017)Vaswani, Shazeer, Parmar, Uszkoreit, Jones, Gomez,
  Kaiser, and Polosukhin]{vaswani2017attention}
Ashish Vaswani, Noam Shazeer, Niki Parmar, Jakob Uszkoreit, Llion Jones,
  Aidan~N Gomez, {\L}ukasz Kaiser, and Illia Polosukhin.
\newblock Attention is all you need.
\newblock In \emph{NeurIPS}, 2017.

\bibitem[Voita et~al.(2019)Voita, Sennrich, and Titov]{voita2019bottom}
Elena Voita, Rico Sennrich, and Ivan Titov.
\newblock The bottom-up evolution of representations in the transformer: A
  study with machine translation and language modeling objectives.
\newblock In \emph{EMNLP-IJCNLP}, 2019.

\bibitem[Wang et~al.(2019)Wang, Singh, Michael, Hill, Levy, and
  Bowman]{wang2018glue}
Alex Wang, Amapreet Singh, Julian Michael, Felix Hill, Omer Levy, and Samuel~R.
  Bowman.
\newblock {GLUE}: A multi-task benchmark and analysis platform for natural
  language understanding.
\newblock In \emph{ICLR}, 2019.

\bibitem[Wang et~al.(2017)Wang, Hamza, and Florian]{wang2017bilateral}
Zhiguo Wang, Wael Hamza, and Radu Florian.
\newblock Bilateral multi-perspective matching for natural language sentences.
\newblock In \emph{IJCAI}, 2017.

\bibitem[West et~al.(2019)West, Holtzman, Buys, and Choi]{west2019bottlesum}
Peter West, Ari Holtzman, Jan Buys, and Yejin Choi.
\newblock Bottlesum: Unsupervised and self-supervised sentence summarization
  using the information bottleneck principle.
\newblock In \emph{EMNLP}, 2019.

\bibitem[White et~al.(2017)White, Rastogi, Duh, and
  Van~Durme]{white2017inference}
Aaron~Steven White, Pushpendre Rastogi, Kevin Duh, and Benjamin Van~Durme.
\newblock Inference is everything: Recasting semantic resources into a unified
  evaluation framework.
\newblock In \emph{IJCNLP}, 2017.

\bibitem[Williams et~al.(2018)Williams, Nangia, and Bowman]{williams2018broad}
Adina Williams, Nikita Nangia, and Samuel Bowman.
\newblock A broad-coverage challenge corpus for sentence understanding through
  inference.
\newblock In \emph{NAACL}, 2018.

\bibitem[Wolf et~al.(2019)Wolf, Debut, Sanh, Chaumond, Delangue, Moi, Cistac,
  Rault, Louf, Funtowicz, and Brew]{Wolf2019HuggingFacesTS}
Thomas Wolf, Lysandre Debut, Victor Sanh, Julien Chaumond, Clement Delangue,
  Anthony Moi, Pierric Cistac, Tim Rault, R{\'e}mi Louf, Morgan Funtowicz, and
  Jamie Brew.
\newblock Huggingface's transformers: State-of-the-art natural language
  processing.
\newblock \emph{arXiv:1910.03771}, 2019.

\bibitem[Yang et~al.(2019)Yang, Dai, Yang, Carbonell, Salakhutdinov, and
  Le]{yang2019xlnet}
Zhilin Yang, Zihang Dai, Yiming Yang, Jaime Carbonell, Russ~R Salakhutdinov,
  and Quoc~V Le.
\newblock Xlnet: Generalized autoregressive pretraining for language
  understanding.
\newblock In \emph{NeurIPS}, 2019.

\bibitem[Zhang et~al.(2017)Zhang, Rudinger, Duh, and
  Van~Durme]{zhang2017ordinal}
Sheng Zhang, Rachel Rudinger, Kevin Duh, and Benjamin Van~Durme.
\newblock Ordinal common-sense inference.
\newblock In \emph{TACL}, 2017.

\bibitem[Zhang et~al.(2021)Zhang, Wu, Katiyar, Weinberger, and
  Artzi]{zhang2020revisiting}
Tianyi Zhang, Felix Wu, Arzoo Katiyar, Kilian~Q Weinberger, and Yoav Artzi.
\newblock Revisiting few-sample bert fine-tuning.
\newblock \emph{ICLR}, 2021.

\bibitem[Zhang et~al.(2015)Zhang, Zhao, and LeCun]{zhang2015character}
Xiang Zhang, Junbo Zhao, and Yann LeCun.
\newblock Character-level convolutional networks for text classification.
\newblock In \emph{NeurIPS}, 2015.

\end{thebibliography}
\bibliographystyle{iclr2021_conference}

\clearpage

\appendix
\section{Experimental Details}\label{app:experiments_details}
\paragraph{Datasets Statistics}
Table~\ref{tab:data-stats} shows the statistics of the datasets used in our experiments.
\begin{table}[H]
    \caption{Datasets used in our experiments.}
    \centering
    \resizebox{0.6\textwidth}{!}{%
    \begin{tabular}{lllllH}
    \toprule 
    \textbf{Dataset}     & \textbf{\#Labels}& \textbf{Train} & \textbf{Val.} & \textbf{Test} & \textbf{Domain} \\
    \midrule 
    \multicolumn{6}{c}{\textbf{Single-Sentence Tasks}} \\ 
    \midrule 
   IMDB & 2 & 20K & 5K & 25K \\ 
   YELP & 5 & 62.5K &  7.8K & 8.7K\\ 
   \midrule 
    \multicolumn{6}{c}{\textbf{Inference Tasks}} \\ 
    \midrule 
    SNLI & 3 & 550K & 10K & 10K \\
    MNLI & 3 & 393K & 9.8K & 9.8K &  misc.\\ 
    RTE &  2 & 2.5K & 0.08K & 3K & news, Wikipedia \\ 
    \midrule 
    \multicolumn{6}{c}{\textbf{Similarity and Paraphrase Tasks}} \\ 
    \midrule
    MRPC & 2 & 3.7K & 0.4K & 1.7K & news\\ 
    STS-B & 1 (Similarity score) & 5.8K & 1.5K & 1.4K & misc.\\ 
    \bottomrule
    \end{tabular}}
    \label{tab:data-stats}\vspace{-1em}
\end{table}

\paragraph{Computing Infrastructure}We run all experiments on one GTX1080Ti GPU with  11 GB of RAM.

\paragraph{VIBERT Architecture} 
The MLP module used to compute the compressed sentence representations (Figure ~\ref{fig:IB}) is a shallow MLP with $768$, $\frac{2304+K}{4}$, $\frac{768+K}{2}$ hidden units with a ReLU non-linearity, where $K$ is the bottleneck size. Following~\citet{alemi2016deep}, we average over 5 posterior samples, i.e., we compute $p(y|x) = \frac{1}{5} \Sigma_{i=1}^5 q_{\phi}(y|z_i)$, where $z^i \sim p_{\theta}(z|x)$.  Similar to~\citet{bowman2016generating}, we use a linear annealing schedule for $\beta$ and set it as $\min(1,~\text{epoch}\times \beta_0)$ in each epoch, where $\beta_0$ is the initial value.

\section{Hyper-parameters}\label{app:hyperparameters}
\paragraph{The GLUE Benchmark Experiment}Results on GLUE benchmark are reported in Table~\ref{tab:glue_results}. We fine-tune all the models for 6 epochs to allow them to converge. We use early stopping for all models by choosing the model performing the best on the validation set with the evaluation criterion of average F1 and accuracy for MRPC, accuracy for RTE, and average Pearson and Spearman correlations for STS-B. For VIBERT, we sweep $\beta$ over $\{10^{-4},10^{-5},10^{-6}\}$ and $K$ over $\{144,192,288,384\}$. For dropout, we use dropping probabilities of $\{0.25,0.45,0.65,0.85\}$. For Mixout, we consider mixout probability of $\{0.1,0.2,0.3,0.4,0.5,0.6,0.7,0.8,0.9\}$. For WD,  we consider weight decay of $\{10^{-6},10^{-5},10^{-4},10^{-3},10^{-2},10^{-1},1\}$.

\paragraph{Varying-resource Experiment} Results on varying sizes of training data are reported in Table~\ref{tab:sampled_results_test}. We fine-tune all models for 25 epochs to allow them to converge. We use early stopping for all models based on the performance on the validation set. We also perform hyper-parameter tuning on the validation set. Since we consider datasets of a different number of training samples, we need to account for a suitable range of bottleneck size and we sweep $K$ over $\{12,18,24,36,48,72,96,144,192,288,384\}$ and $\beta$ over $\{10^{-4},10^{-5}\}$. For dropout, we consider dropping probabilities of $\{0.25,0.45,0.65,0.85\}$. For Mixout, we consider mixout probability of $\{0.1,0.2,0.3,0.4,0.5,0.6,0.7,0.8,0.9\}$. For WD, we consider weight decay of $\{10^{-6},10^{-5},10^{-4},10^{-3},10^{-2},10^{-1},1\}$.

\paragraph{Ablation Experiment} Ablation results are shown in Table~\ref{tab:glue_results_ablation}. For VIBERT ($\beta$=0), we sweep $K$ over the same range of values as VIBERT, i.e., $\{144,192,288,384\}$

\section{Mapping} \label{app:transfer}
We train all models on SNLI or MNLI datasets and evaluate their performance on other target datasets. The SNLI and MNLI datasets contain three labels of contradiction, neutral, and entailment. However, some of the considered target datasets have only two labels, such as DPR or SciTail. When the target dataset has two labels of \emph{entailed} and \emph{not-entailed}, as in DPR, we consider the predicted contradiction and neutral labels as the not-entailed label.  In the case the target dataset has two labels of \emph{entailment} and \emph{neutral}, as in SciTail, we consider the predicted contradiction label as neutral.

\end{document}